\documentclass[letterpaper,10pt,conference]{IEEEtran}

\IEEEoverridecommandlockouts                              

\usepackage{graphics} 
\usepackage{graphicx}
\usepackage{array}
\usepackage{float}
\usepackage{subfigure}

\usepackage{url}
\usepackage{hyperref}
\usepackage{amsmath}
\usepackage{algpseudocode}
\usepackage{amssymb}
\usepackage{enumitem}
\usepackage{algorithm}
\usepackage{hhline}

\begin{document}

\title{\LARGE \bf
Algae Detection Using Computer Vision and Deep Learning
}

\author{{Arabinda Samantaray$^{1}$, Baijian Yang$^{2}$, J. Eric Dietz$^{2}$, and Byung-Cheol Min$^{1}$ 
\thanks{$^{1}$ A. Samantaray and B.-C. Min are with  SMART Lab, Department of Computer and Information Technology, Purdue University, West Lafayette, IN 47907, USA. e-mails: \tt{samantar@purdue.edu,minb@purdue.edu}} 

\thanks{$^{2}$ B. Yang and J. E. Dietz are with Department of Computer and Information Technology, Purdue University, West Lafayette, IN 47907, USA. e-mails: \tt{byang@purdue.edu,jedietz@purdue.edu}}}
}

\maketitle

\begin{abstract}
A disconcerting ramification of water pollution caused by burgeoning populations, rapid industrialization and modernization of agriculture, has been the exponential increase in the incidence of algal growth across the globe. Harmful algal blooms (HABs) have devastated fisheries, contaminated drinking water and killed livestock, resulting in economic losses to the tune of millions of dollars. Therefore, it is important to constantly monitor water bodies and identify any algae build-up so that prompt action against its accumulation can be taken and the harmful consequences can be avoided. In this paper, we propose a computer vision system based on deep learning for algae monitoring. The proposed system is fast, accurate and cheap, and it can be installed on any robotic platforms such as USVs and UAVs for autonomous algae monitoring. The experimental results demonstrate that the proposed system can detect algae in distinct environments regardless of the underlying hardware with high accuracy and in real time. 

\end{abstract}


\section{INTRODUCTION}
\label{sec:int}
Algae are primarily aquatic, uni or multi-cellular organisms that contain chlorophyll \cite{Algae}. In a healthy aquatic environment, algae play the role of primary producers and are critical in preserving the food chain. Algae also benefit humans by reducing the level of greenhouse gases in the atmosphere by fixing large quantities of $CO_2$ in oceans \cite{warming}, serving as a source of energy in the form of bio fuels \cite{menetrez2012overview} and acting as a cheap but highly effective means for waste-water treatment \cite{fallowfield1985treatment}.

Under congenial ecological conditions, however, the rate of algae proliferation can increase exponentially, and large algal colonies, sometimes covering an area of many square kilo-meters, can be formed. Such colonies are known as harmful algal blooms (HABs). They have been found to be responsible for releasing paralytic, neurotoxic, diarrhetic, amnesic and azaspiracid  toxins causing shellfish poisoning \cite{Anderson2009} in bodies of water, leading to the deaths of fishes, sea mammals, birds and even humans \cite{SBhat2004}\cite{Okaichi}. 

\begin{figure}[t]
\centering
\includegraphics[width=0.90\linewidth]{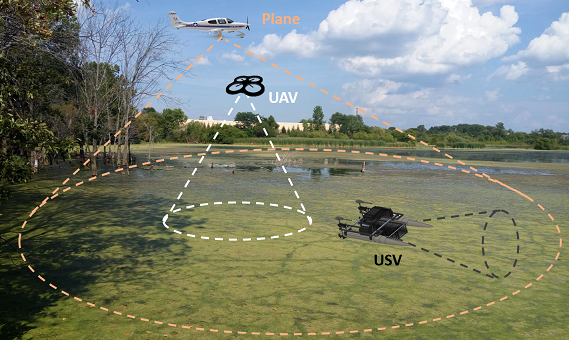}
\caption{Concept of a real-time algae monitoring using a mobile platform. Our proposed computer system can be installed on any robotic platforms such as USVs and UAVs.}
\label{fig_concept}
\end{figure}

In addition to releasing toxins, prolonged periods of algal cover on water surfaces reduces sunlight penetration, thus preventing underwater aquatic flora from conducting photo-synthesis and negatively affecting the aquatic ecosystem. Subsequently, when HABs decay, they consume large quantities of oxygen, causing anoxia and death of marine mammals, fishes and crustaceans among others \cite{Anderson2009}. 

The destruction of aquatic environments by HABs has a significant negative impact on quality of life for many local communities that depend on these  water bodies for potable water, recreation and a flourishing fishing and tourism industry \cite{algaecost}.

Hence, it is extremely important that ponds, lakes and rivers are constantly monitored and prompt action is taken against any abnormal algae buildup. However, current methodologies for algae monitoring require significant manpower, making them economically unfeasible (e.g., it was projected that 
a budget of \$3.5 million was required to monitor the 100 largest lakes in Oklahoma once a month \cite{hambright2015remote}). Alternatively, methods may depend upon the hardware available, which can be in the form of drones \cite{jung}, robotic fish \cite{Yu}, underwater cameras \cite{tan2014framework} or satellites \cite{carvalho2010satellite}, which limits their applicability in varying environmental, topological and socio-economic conditions.

The primary contributions of our research include the following:
\begin{itemize}
\item Development of a computer vision algorithm based on deep learning that can detect and locate algae accurately in water bodies, irrespective of the variations in camera parameters, environmental conditions, orientation of the captured image or the presence of significant background \textit{clutter} in the form of trees, vegetation, buildings etc.
\item Our proposed system can be installed on mobile platforms including unmanned aerial vehicles (UAVs), unmanned surface vehicles (USVs), and airplanes. when they are used for algae monitoring, because the system can detect algae in real time.
\item The proposed computer vision system is cheap as it does not require any specialized hardware.
\end{itemize}


This paper is organized as follows. In Section \ref{sec:rel}, we describe the current algae monitoring systems and their shortcomings.  We introduce the proposed algae detection system using computer vision and deep learning in Section \ref{sec:met}. In Section \ref{sec:eval}, we present the evaluation procedure and the results obtained, followed by conclusions and future works in Section \ref{sec:con}.

\section{RELATED WORK}
\label{sec:rel}
Eutrophication by agricultural runoff \cite{algaeblooms}, dumping of untreated industrial and household effluents \cite{wastes} and transportation of foreign algae species in ballast water \cite{Hallegraeff}\cite{Hallegraeff1}, coupled with natural phenomena such as storms, tsunamis, currents \cite{Kevin} and global warming \cite{Louis}, have been cited as the root cause for the rising incidence of HABs in water bodies across the globe. This rising incidence has increased the relevance of effective algae monitoring methodologies in the modern world. 

Algae monitoring techniques can be roughly classified into 3 categories: \textit{in}-\textit{situ} sampling, computer vision-based techniques and hyperspectral remote sensing using satellite or aircraft. 

\subsection{In-situ Sampling} 
\textit{In}-\textit{situ} sampling is done by performing on-site sampling and transporting those samples to laboratories for further evaluation. Although onsite sampling is done at regular intervals, this methodology is extremely time and labor intensive. Also, the possibility of contaminated samples negatively affecting observations is high \cite{glasgow2004real}.

\subsection{Computer Vision based Techniques}
The distinctive green or greenish-blue color, characteristic of algae has been used to develop computer vision-based algae monitoring systems. However, such traditional computer vision pipelines do not have high repeatability because they are significantly dependent on the effectiveness of the feature detectors or the segmentation procedure, which can be rendered ineffective by various environmental conditions such as fluctuating illumination, occlusion or the presence of comparable objects in the background \cite{lecun1998gradient}. 

The use of color-based segmentation, species specific features and an underwater camera in study \cite{tan2014framework} implies that the algae monitoring system cannot be applied on a different hardware platform such as an UAV, USV, or  smartphone and is susceptible to occlusion and variations in illumination, which are regular occurrences in an outdoor environment. Similarly, the novel approach of combining the use of a smartphone camera and the inertial sensors present on a robotic fish to detect the shoreline and hence perform image segmentation to detect algal blooms as presented in the paper \cite{Yu} is optimized for their specific hardware platform (i.e., robotic fish) and is susceptible to variable illumination, occlusion and shadows cast by surrounding vegetation.

Similar limitations can also be observed in the work of \cite{jung}, which made use of a local binary pattern (LBP) texture detector to detect algae. However, in their work, they have only indicated the success of their methodology by using a UAV platform 
and have not described their system's performance in case the images were taken from  different orientation or higher elevation. Also,they have not presented any results describing the impact that the presence of comparable objects such as trees, plants and seaweeds in the image's background would have on their algae monitoring system, considering that their detector is only trained on \textit{iconic} images of algae, grass and water.  
 
Even in paper \cite{kumar2017deep}, the authors only considered images that were taken from the ground and did not describe the speed with which their vision system could detect algae. Hence, it is difficult to ascertain whether this method could be used from mobile platforms such as UAVs, USVs and airplanes, which operate in different environments.

\subsection{Satellite Remote Sensing}
The use of satellite-based remote sensing for algae monitoring makes use of the increased diffused reflectance caused by the presence of algal pigment in a body of water \cite{carvalho2010satellite}. Different categories of algorithms such as reflectance classification algorithms, reflectance band-ratio algorithms and spectral band difference algorithms take the spectral data as input to detect the presence of algae in bodies of water \cite{blondeau2014review}.

However, although these algorithms have been successful in monitoring algal blooms in the open ocean, they have been unproductive in coastal water and in bodies of water with significant human activity because the reflectance spectrum becomes distorted in the presence of organic material and suspended particles. Also, issues such as unavailability of real-time data, irregular site revisit times, low resolution of publicly available satellite products such as LANDSAT or MODIS ($>30m$) and exorbitant costs of proprietary systems such as QuickBird \cite{flynn2014remote} make the use of satellite imagery for a general purpose algae monitoring system difficult.

\begin{figure}[!t]
    \centering
\subfigure[Ground view]{\label{fig:temp}\includegraphics[width=0.24\textwidth]{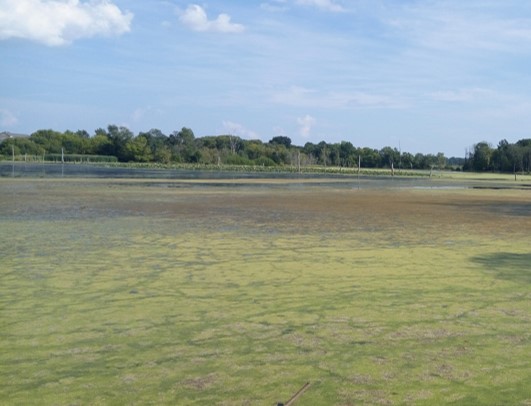}}
\subfigure[Aerial view]{\label{fig:temp}\includegraphics[width=0.24\textwidth]{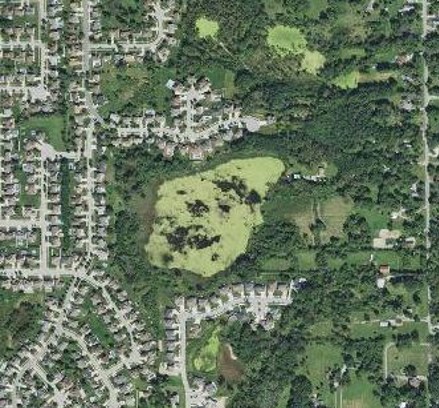}}
    \caption{Different images of algae filled pond taken from a ground view and an aerial view.}
\label{fig_algae}
\end{figure}

\begin{figure*}[!t]
\centering
\includegraphics[width=0.95\textwidth]{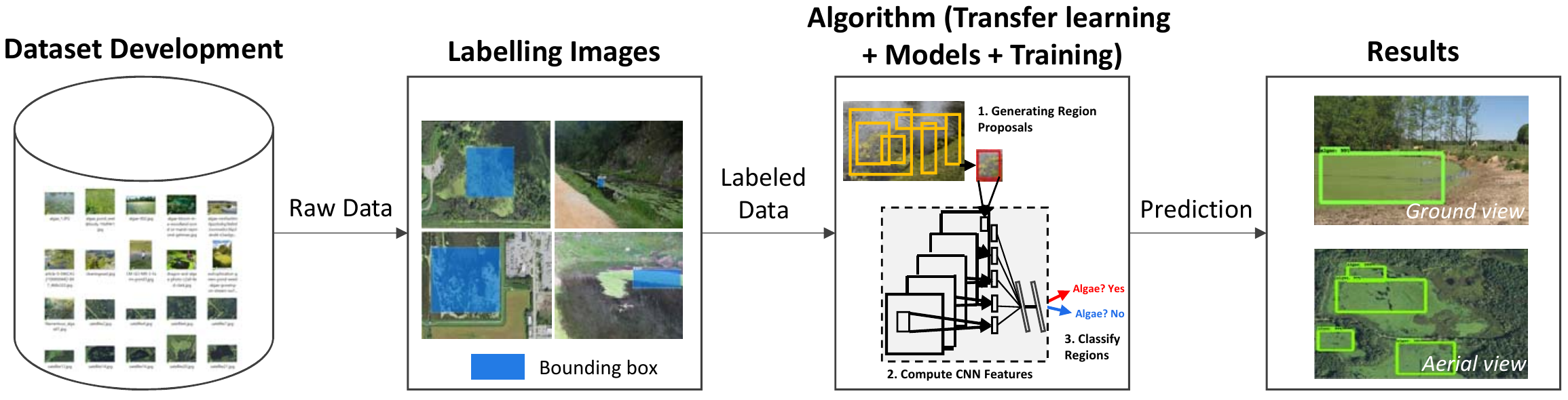}
\caption{The procedure undertaken to develop the proposed computer vision system for algae detection.}
\label{fig_met}
\end{figure*}

Hence, an economically feasible algae monitoring system that is robust to changes in image parameters (e.g., image size, resolution, orientation), can be easily calibrated depending on the environmental conditions and algal species under consideration and is able to work from a wide variety of platforms (e.g., UAVs, USVs, airplane, and even smartphones) would significantly facilitate administrators and civilians in monitoring bodies of water for algal blooms. The development of such a computer vision system is the major contribution of this paper.

\section{Methodology}
\label{sec:met}
Current computer vision-based algae monitoring techniques can be classified into three categories. The first comprises methods that use UAVs, satellites or airplanes for algae monitoring. The second class of monitoring techniques makes use of water based robots, USVs and ships. A third approach that has been proposed is to build a community-based algae monitoring system composed of citizen scientists using ground-based cameras such as smartphones \cite{kumar2017deep}. However, all these techniques capture images from different orientations and use computer vision pipelines that exploit the advantages provided by their respective hardware platforms, making it highly improbable that the vision system developed for aquatic platforms could work with images taken from the aerial view or vice-versa because the images would be significantly different from each other as shown in Figure \ref{fig_algae}.

However, using only one kind of platform severely restricts our ability to monitor bodies of water of different sizes effectively. A ground-based algae monitoring system would be incapable of detecting the growth of algal blooms at the center of large lake, while aerial or aquatic robot platform based   methods would restrict algae monitoring capabilities to only communities that possess that particular hardware. Hence, a computer vision system that enables algae monitoring to be executed by different platforms, as depicted in Figure \ref{fig_concept}, would democratize the process of algae monitoring and make it available to communities all across the globe.

The following subsections describe the steps shown in Figure \ref{fig_met}, which reflects the steps we have undertaken to develop our computer vision system.

\subsection{Dataset Development}
The first step in the application of machine learning algorithms is the preparation of a dataset. Because there is no publicly available dataset containing images of algae in an outdoor environment, a new benchmark dataset must be developed. Each image in the dataset should be labeled with annotation software to generate files containing the coordinates of the bounding boxes, indicating the location of algae in the image. To ensure that the complexity of the images in the dataset is equivalent to that of the outdoor environment, it is recommended that a significant number of background objects such as trees, grass, plants, roads and buildings be present in the images, rather than just having the region of interest, in this case algae, as the main part of the image. 

Subsequently, the images and their corresponding annotations should be randomly assigned to training, validation and testing datasets. Typically, we can assign 70\% of images to the training set, 20\% of images to the validation set and 10\% to the test set.

\subsection{Object Detection Algorithm }
Initially, the general lack of algae images on the Internet made it inconceivable to apply deep learning algorithms due to the unavailability of a large amounts of annotated data,akin to conventional datasets such as COCO \cite{lin2014microsoft} or PASCAL-VOC \cite{everingham2010pascal}.
However, a new paradigm of machine learning known as \textit{Transfer Learning} enabled us to overcame the limitations posed by the size of our dataset.

\subsubsection{Transfer Learning}
Transfer learning is a new paradigm in machine learning that enables us to use different domains, distributions,  and tasks in training and testing datasets \cite{pan2010survey}. This implies that, Convolutional Neural Network (CNN), which has been trained on a large labeled dataset from a different domain can be applied for feature detection in a completely new domain. This is made possible by the fact that the lower layers of the network are capable of detecting general features such as edges, blobs etc., that comprise an image, while higher layers capture domain specific features. This enables a pre-trained network to be learn from a much smaller dataset, to recognize a completely new class of objects, as only the latter layers of the network have to be trained \cite{shin2016deep}.

\subsubsection{Deep Learning Models}

An algae monitoring system, that conceivably could be used from mobile platforms such as USVs, UAVs, airplanes etc., has to detect and locate algae at near real time speeds with high accuracy. As Faster R-CNN, Single Shot Detector (SSD), and Region-based Fully Convolutional Networks (R-FCN) models have shown near real-time object detection on conventional datasets such as COCO and PASCAL-VOC with very high accuracy, they are applicable to our envisaged scenario. The working of these 3 models is described in the following subsections.


\begin{itemize}

\item \textit{Faster R-CNN}:
The Faster R-CNN \cite{ren2017faster} was developed to remove the limitations of R-CNN \cite{girshick2014rich} and Fast R-CNN \cite{girshick2015fast} models. It makes use of a region proposal network to generate box proposals from an intermediate feature map. These box proposals are then applied on the same feature map to extract proposal regions, on which class predictions and bounding box improvements are made.

\item \textit{R-FCN:}
R-FCN network has a shared convolution architecture that makes the use of position sensitive score maps to compromise between translation invariance for image classification and translation variance for object detection \cite{dai2016r}. 

\item \textit{Single Shot Detector:}
The single shot multibox detector was recently developed in \cite{liu2016ssd}. It is a single layer feed forward convolution network that makes the use of pre-defined anchor boxes, to train the network and make predictions about the class within the anchor box and the offset by which the anchor needs to be shifted to fit the ground-truth box. 

\end{itemize}

\subsubsection{Learning from Custom Dataset}

\begin{figure}[t]
    \centering
\subfigure[Results]{\label{fig:temp}\includegraphics[width=0.24\textwidth]{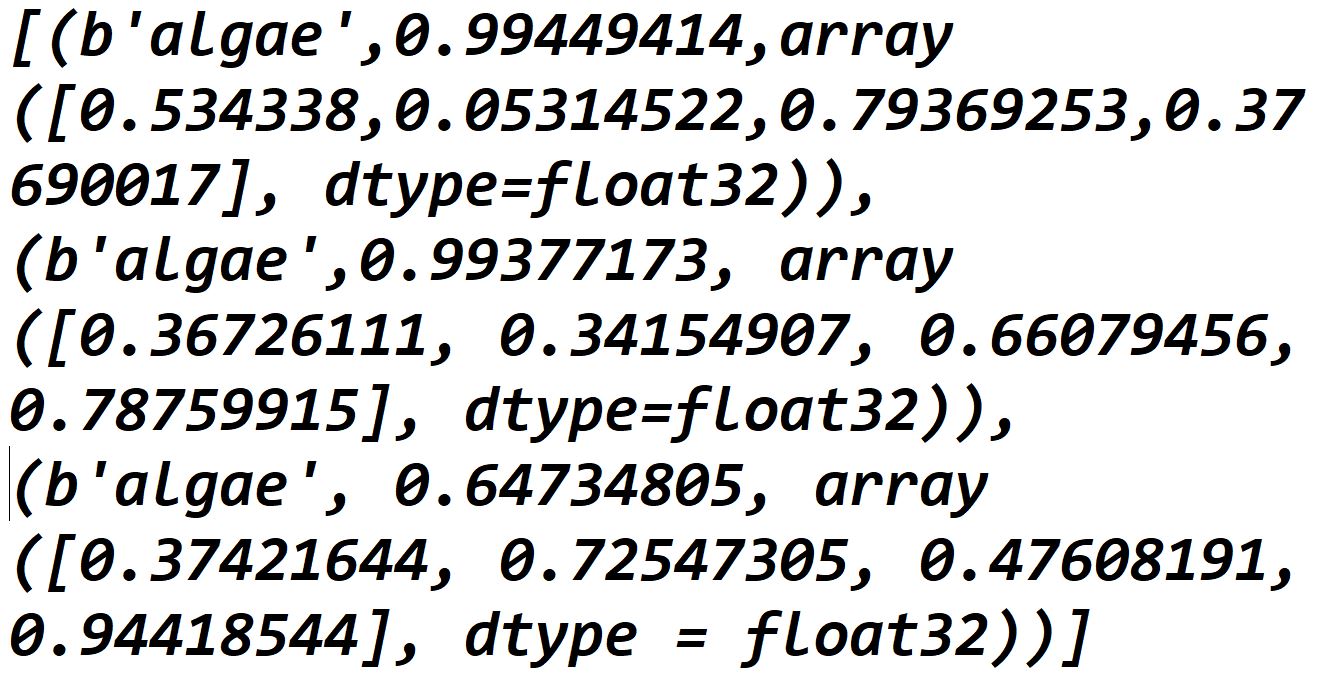}}
\subfigure[Resultant Image]{\label{fig:temp}\includegraphics[width=0.24\textwidth]{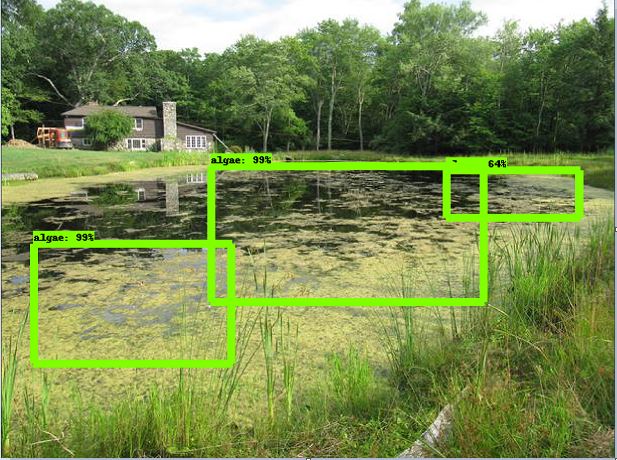}}
    \caption{Results generated by the model and visualizing them on the respective image.}
\label{fig_results}
\end{figure}

The procedures followed to enable the chosen deep learning models to detect algae are as follows:

\begin{enumerate}[label={\alph*})]
\item \textit{Convert data:} 
The input data, i.e., the images and bounding box coordinates, need to be converted into appropriate formats depending upon how the data is being ingested and the deep learning framework being utilized. In case of Tensorflow \cite{ObjectDetection}, the input data should be converted into \textit{tfrecords} while for MXNET \cite{mxnet}, \textit{recordIO} file should be used. This ensures faster processing as compared to when the data is read directly from disk.

\item \textit{Data Augmentation:} To mitigate the effects of having a small dataset and to replicate external environmental conditions such as variable illumination, fluctuating contrast, and blurring, the dataset should be augmented by applying transformations such as randomly changing the brightness, contrast, hue, color, and saturation.

\item \textit{Label Maps:} 
Machine learning algorithms cannot work with categorical features (e.g., class labels for object of interest in images), and hence categorical features should be related with a numerical value. This is usually done by using a label map. Since our dataset annotations belong to only one class (i.e., algae), in our label map, the algae class is related to the ID 1.

\item \textit{Fine tune hyperparameters in configuration files:} 
Pre-trained neural networks are made available with 2 components, which are the weights of the model and the configuration file that determines the meta-architecture of the model, feature extractor present in the model, training parameters and the evaluation metrics. To customize the pre-trained networks for our dataset, a decaying learning rate of 10\% every 5000 steps is applied, and we change the final layer to reflect that there is only one class of objects in our dataset. 

\item \textit{Begin training and monitor evaluation metrics:} 
After customizing the configuration files, the training of the neural network models can be initialized, which would fine-tune the weights in the latter layers of the neural networks enabling them to detect algae. During the training of the model, after a specified number of steps, an intermediate trained model would be stored and evaluated on the validation dataset. Observing the different metric values such as training loss, mean average precision (mAP) on the validation dataset enables us to infer:
\begin{itemize}
\item When the model has stopped learning, so that training can be stopped
\item Whether subsequent iterations of training should be conducted, by optimizing hyperparameters or using a larger dataset
\end{itemize}

%
%

\renewcommand{\algorithmicrequire}{\textbf{Input:}}  
\renewcommand{\algorithmicensure}{\textbf{Output:}} 

\begin{algorithm}[t]
  \caption{Training a neural network to detect algae}
  \label{alg::conjugateGradient}
  \begin{algorithmic}[1]
    \Require
      A file containing pre-trained weights of the model;
      Label map;
      Configuration file for the pre-trained model;
    \Ensure
      A file containing the weights of the newly trained model    
    \Repeat
      \State Prepare an annotated dataset and split it into training, validation and testing dataset
      \State Convert the dataset annotations into appropriate input format 
      \State Fine tune the hyperparameters of the neural network
      \State Monitor the training loss and mean average precision on validation dataset
      \State If mAP graph converges stop training and observe the final 	 validation mAP
    \Until{validation mAP $>$ satisfactory mAP}
    \State Obtain the mAP of the trained network on the test dataset
    \State Deploy the model into production
     \State Set a confidence threshold and visualize the results in the image
  \end{algorithmic}
  \label{alg-ten}
\end{algorithm}

\item \textit{Evaluating the trained model on the test dataset and deploying into production:}
The trained model would be evaluated on the test set to examine its classification and detection accuracy  on a completely new set of data, allowing us to infer whether the chosen model is applicable to be used as an algae monitoring system.

\item \textit{Visualizing results generated by neural network:}
For each frame/image the neural network populates the following arrays:
\begin{itemize}
\item Boxes -- this array contains the normalized coordinates for each predicted bounding box.
\item Scores -- this array contains the confidence scores for each of the predicted boxes.
\item Classes -- this array contains the class label for each of the predicted boxes.
\item Number of detections -- This array contains the value regarding total number of detections made per image.
\end{itemize}
From these individual arrays, we create a list of all the predicted bounding boxes having a confidence score of higher than 50\%. Each list item contains the class label, normalized box coordinates and the confidence scores for each bounding box as shown in Figure \ref{fig_results}.

By applying the following equations for each bounding box, we convert the normalized coordinates into image coordinates,
\begin{equation}
Coordinate_k = Box_{i}^{j}\cdot ImageWidth
\end{equation}
where $k\in$~(left,right,top,bottom), $i$ is an index of boxes, $j \in(0,1,2,3)$, and $ImageWidth$ is a width of the image. Subsequently, these image coordinates can then be used to visualize the results of the predicted boxes as shown in Figure \ref{fig_results}.

%
%

%

\end{enumerate}

Algorithm \ref{alg-ten} summarizes all the steps from (a)--(g), involved in developing deep learning models to detect algae.


\section{Evaluation and Results}
\label{sec:eval}

\subsection{Preparation for Evaluation}
Since our objective was to develop a computer vision system that can detect and locate algae in water bodies at a fast speed, we focused on the following three  different evaluation metrics:
\begin{itemize}
\item Precision and Recall -- to evaluate the accuracy of our system in detecting whether a given water body contains algae or not
\item Mean Average Precision -- to evaluate how accurately our system can locate an algal bloom in a water body
\item Speed -- to evaluate the speed at which each neural network detects algae, so as to validate our approach's appropriateness to be used in mobile platforms such as USVs, UAVs, airplanes etc.
\end{itemize}

\subsubsection{Dataset Development}

For purpose of this research, we developed a dataset containing images of algae in pools, lakes, ponds etc., taken from ground and aerial vehicles. We also collected some images from aerial vehicles of water bodies not containing algal blooms. The dataset we developed had 4 categories:

\begin{itemize}
\item Training dataset -- this dataset was used to train each of our neural networks
\item Validation dataset -- this data was used to validate the performance of our training, by evaluating the intermediate neural networks. Based upon the results on the validation set, we decided whether to continue training, fine-tune hyper-parameters or stop training.
\item Test set for classification -- to ensure that our trained neural networks were only detecting water bodies containing algae, and not water bodies in general, we developed a testing dataset containing images of water bodies having algae (52 images) and not having algae (48 images).
\item Test set for detection -- on this dataset, we had images containing ground truth boxes around algal bloom patches and the mAP of each network was calculated on this dataset
\end{itemize}

The division of the entire dataset into training, validation, and testing dataset is presented in Table \ref{tab-dis}.

\begin{table}[!t]
\renewcommand{\arraystretch}{1.3}
\caption{Distribution of images across the training, validation, and test sets (D.= Detection)}
\centering
\begin{tabular}{|c|c|c|c|c|}
\hline
& \bfseries  Training & \bfseries Validation & \bfseries Testing (D.)\\
\hline
Ground Images & 277 & 79 & 41\\
\hline
Aerial Images & 150 & 43 & 20\\
\hline
\end{tabular}
\label{tab-dis}
\end{table}

\subsubsection{Model Training Environment}
We utilized \textit{Tensorflow Object Detection} API to train our chosen models to detect algae. \textit{Tensorflow Object Detection} API is an open source framework based on the tensorflow library, and it provides a well structured environment for developing, training, testing, and deploying deep learning models. 

\subsubsection{Hardware for Evaluation}
For evaluating our proposed computer vision system, we use a HP Pavilion laptop, having a Intel(R) Core(TM) i7-6500U CPU and NVIDIA GeFORCE 940MX GPU.

\subsection{Results and Analysis}

In this section, we will describe the results obtained by the 3 different neural networks on each evaluation criterion:

\subsubsection{Classification}

A reliable algae monitoring system should have high \textit{Recall}( if algae is present, then the system should detect it) and high \textit{Precision}(if the presence of algae is predicted, then algae should be present in the water body). To evaluate the competence of our system in detecting algae, we performed a binary classification between two sets of images, one having water bodies containing algae and others having water bodies not containing algae. The trained version of each model was tested on the aforementioned images and their results can be seen from the Table \ref{tab-all-results} (See column 2--4). Since there were only two class of images, we can observe that SSD performed very poorly and its ability to detect algae was akin to that of random selection which implies that it is not at all a suitable model for algae detection. However, while Faster R-CNN and R-FCN had nearly similar \textit{Precision} values, the high \textit{Recall} value of R-FCN indicates that it is highly robust and nearly always detects an algae bloom if it is actually present.

\begin{table*} [t]
\renewcommand{\arraystretch}{1.5}
\caption{Evaluating the accuracy of classification (Accuracy, Recall, and Precision), the accuracy of detection (Validation mAP and Test mAP), and the speed of the detection (FPS (GPU) and FPS (CPU)) of each model (Valid. = Validation)}
\centering
\begin{tabular}{|c||c|c|c||c|c||c|c|}
\hline
 & \bfseries Accuracy & \bfseries ~~Precision~~  & \bfseries Recall & \bfseries Valid. mAP & \bfseries Test mAP & \bfseries FPS (GPU) & \bfseries FPS (CPU) \\
\hline\hline
Faster R-CNN  & 72\% & 74\% &71.15\%  & 38.68\% & 23.46\% & 2.83 & 0.62\\
\hline
R-FCN  & 82\% & 78.33\%  & 90.38\%  & 38.14\% & 21.44\% & 2.72 & 0.58\\
\hline
SSD  & 50\%  & 52.08\% & 48.07\%  & 23.72\% & 17.29\% & 19.37 & 5.16\\
\hline
\end{tabular}
\label{tab-all-results}
\end{table*}

\subsubsection{Detection}

The second objective of our algae monitoring system was to locate the exact position of algal blooms in an image. Accurate algae localization on an image would enable us to generate precise world coordinates of where the algae blooms are present, by applying \textit{Camera Calibration}. The availability of such fine-grained algae monitoring would be a great asset to the local administrators in their battle against the algae menace. To evaluate the algae detection accuracy of the 3 trained models we used the mean average precision (mAP) metric. The mAP is obtained by integrating the precision recall curve \cite{tangobject},
\[  mAP = \int_{0}^{1} p(x) dx \tag{1} \] where \textit{p(x)} 
is the precision-recall curve. To determine the precision recall curve, the true positive and false positive values of the prediction can be computed by using the Intersection over Union (IoU) criterion,

\[ \textit{IOU}=\frac{Box_{pred} \cap Box_{gt}}{Box_{pred} \cup Box_{gt}} \tag{2} \]
where \(Box_{pred} \)
and \(Box_{gt}\)
are the areas included in the predicted and ground truth bounding box, respectively. Then, a threshold for IoU is designated (e.g., 0.5), and if the IoU exceeds
the threshold, the detection is marked as correct detection.
Multiple detections of the same object are considered as one correct detection, while others are considered  as false detections.
Post, obtaining the true positive and false positive values, a precision-recall curve is generated, based on which the mAP can be calculated.

\begin{figure*}[h]
    \centering
\subfigure[Algae detection by R-CNN]{\label{fig:temp}\includegraphics[width=0.9\textwidth]{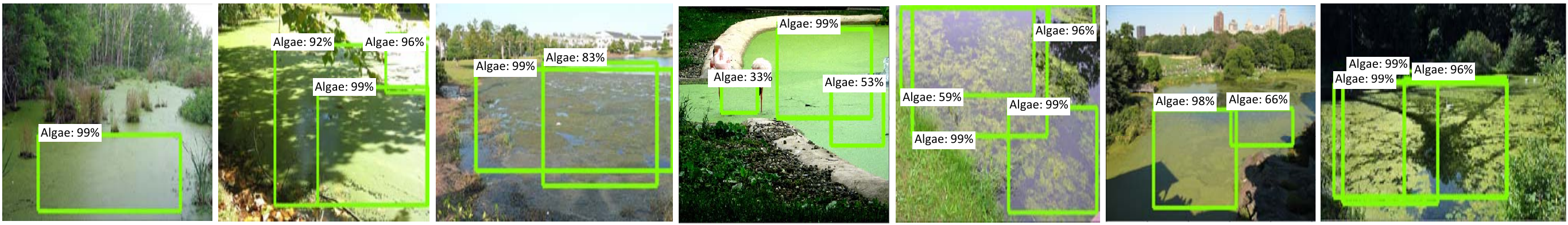}}
\hskip 0.1truein
\subfigure[Algae detection by R-FCN]{\label{fig:temp}\includegraphics[width=0.9\textwidth,height=0.1\textheight]{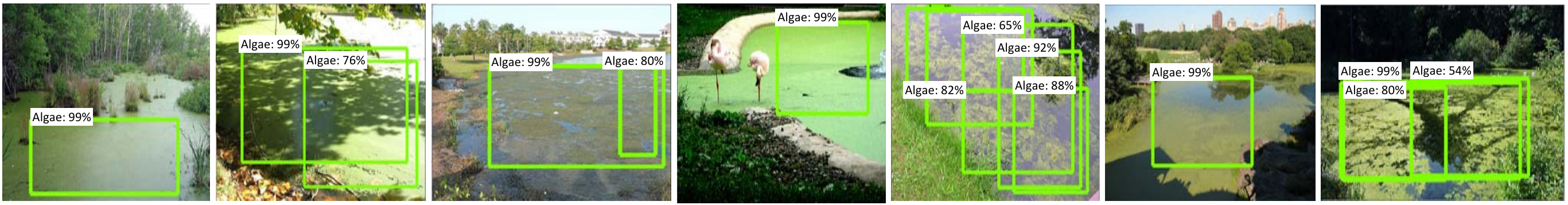}}
\hskip 0.1truein
\subfigure[Algae detection by SSD]{\label{fig:temp}\includegraphics[width=0.9\textwidth,height=0.1\textheight]{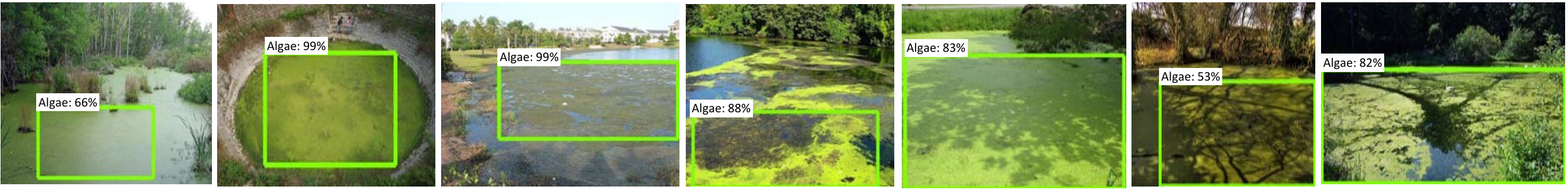}}
    \caption{Images that show detection results for each model from a ground view drawn in green.}
\label{fig_res_gro}
\end{figure*}

\begin{figure*}[!h]
    \centering
\subfigure[Algae detection by R-CNN]{\label{fig:temp}\includegraphics[width=0.9\textwidth]{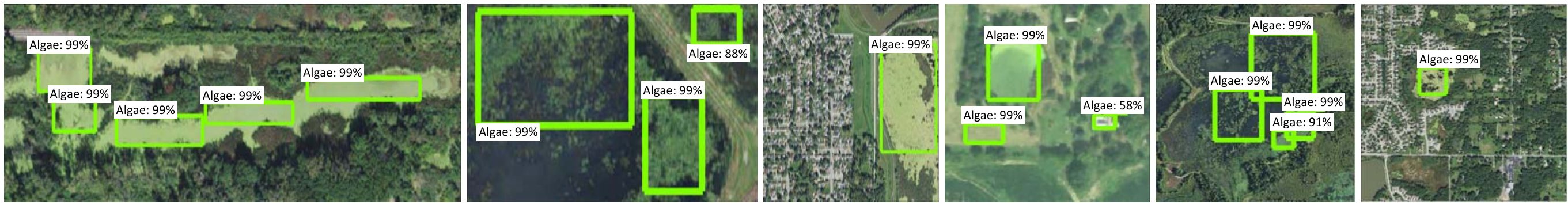}}
\hskip 0.1truein
\subfigure[Algae detection by R-FCN]{\label{fig:temp}\includegraphics[width=0.9\textwidth,height=0.1\textheight]{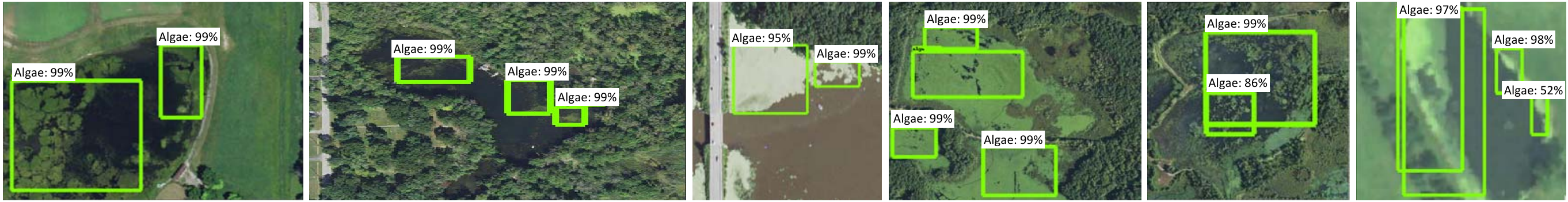}}
\hskip 0.1truein
\subfigure[Algae detection by SSD]{\label{fig:temp}\includegraphics[width=0.9\textwidth,height=0.1\textheight]{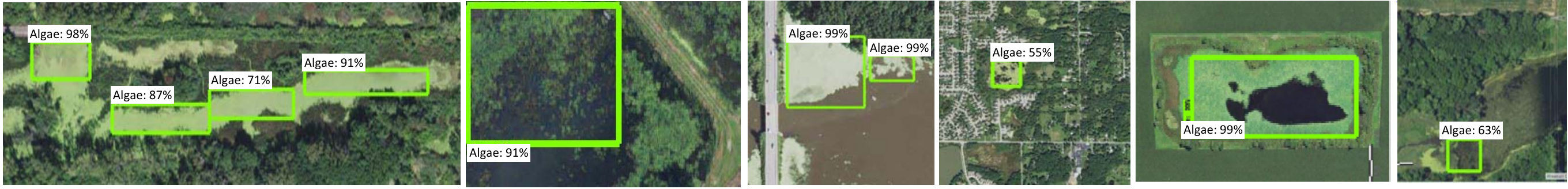}}
    \caption{Images that show detection results for each model from an aerial view drawn in green.}
\label{fig_res_aer}
\end{figure*}

The mAP values obtained by evaluating all the 3 trained models on the validation and test dataset are presented in Table \ref{tab-all-results} (See column 5 and 6). From the table, we can observe that all the 3 models show reasonably acceptable accuracy in algae bloom localization. In particular, it reveals that both Faster R-CNN and R-FCN have higher detection accuracy than SSD on both the test and validation dataset. Also, we can observe that the mAP on the test data set is lower than the mAP on the validation dataset. We believe, this resulted from the fact that we had chosen the most complex images in our entire dataset with significant amounts of background \textit{clutter} (i.e., trees, buildings, roads etc.,) to be a part of our test dataset. Nevertheless, despite the complexity inherent to our test dataset we observe from Figure \ref{fig_res_gro} and Figure \ref{fig_res_aer} that algal blooms are well detected regardless of orientation and location from where the image was taken. 

\subsubsection{Speed of Detection}
Since we envision that our algae monitoring system can be applied to fast moving mobile platforms such as UAV, USV, airplanes etc., it is necessary that our system can perform algae detection and localization in real time. The results regarding the speed with which each model can detect algae in an input image is shown in Table \ref{tab-all-results} (See the last two columns). In contrast to the classification and detection accuracy, SSD outperforms the other two models in terms of a detection speed both when using a CPU and GPU. However, the detection speeds for each model, particularly when implemented on a GPU, are satisfactory for being used as algal monitoring systems as algal blooms grow in static or slow moving water bodies.

\subsection{Discussion}
\label{sec:disc}
\begin{figure}[!h]
    \centering
\subfigure[Incorrect detection of surrounding vegetation as algae]{\label{fig:temp}\includegraphics[width=0.45\textwidth]{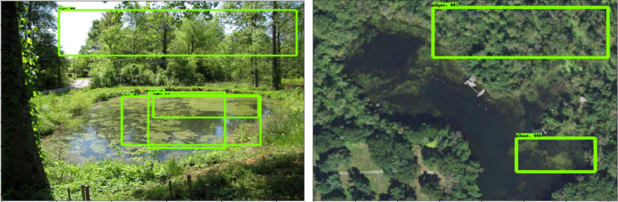}}
\hskip 0.1truein
\subfigure[Inabilty to detect algae from ground and aerial view]{\label{fig:temp}\includegraphics[width=0.45\textwidth]{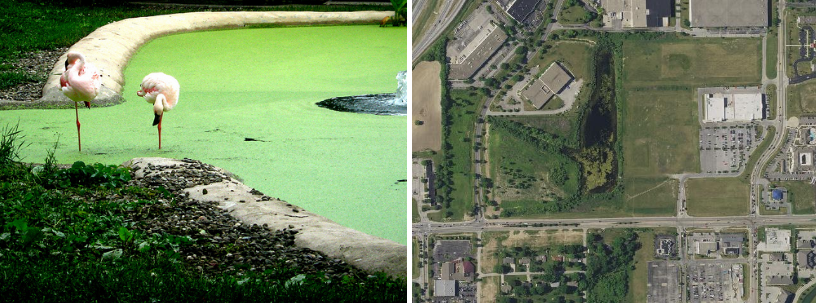}}
    \caption{Images that show incorrect or inability to detect algae.}
\label{fig_err}
\end{figure}

Despite using state-of-art object detection models, we observed that there were a few occasions in which either algae was not detected or the surrounding vegetation was detected as algae, which could be a topic for further study. Some examples are presented in Figure \ref{fig_err}. The most likely cause for such incidences is that, the only defining characteristic of algae is its green color which is the most common color found in an outdoor environment. However, a larger dataset enabling the neural network to learn more intricate features would be able to detect algae with more consistency.



\section{Conclusion and Future Work}
\label{sec:con}
In this work, our objective was to develop a computer vision system that can detect and locate algal blooms in real time, from cameras placed on aerial, aquatic or ground based platforms. In order to attain our objective we compared state-of-art object detectors such as Faster R-CNN, R-FCN, and SSD.Our final conclusion was that an algae monitoring system based on the R-FCN model would be highly robust, accurate and fast to enable effective, real time algae monitoring. We expect that such a  system will significantly facilitate researchers, local administrators and civilians in monitoring water bodies and promptly curbing any excessive algal growth.

Future works will be focused on improving the current performances by developing a larger dataset and implementing field tests.  In addition, we will use this computer vision system to facilitate the working of a multi-robot ecosystem composed of autonomous UAVs and USVs which would be responsible for monitoring water bodies and removal of HABs as shown in Figure \ref{fig_concept}.

\section{Acknowledgements}
The author would like to acknowledge the help and support provided by the SMART Lab members, notably Wonse Jo and Dr. Ramviyas Parasuraman in the development of this computer vision system.
\bibliographystyle{IEEEtran}
\bibliography{references}

\end{document}